\documentclass[letterpaper, 10 pt, conference]{ieeeconf}  %
\IEEEoverridecommandlockouts                              %
\usepackage{graphics} 
\usepackage{graphicx} 
\usepackage{mathptmx} 
\usepackage{times} 
\usepackage{amsmath} 
\usepackage{amssymb}  
\usepackage{color}
\usepackage{float}
\usepackage{stmaryrd}
\usepackage{setspace}
\usepackage{multicol}
\usepackage{hyperref} 

\title{\LARGE \bf  Finding a Landing Site on an Urban Area: A Multi-Resolution Probabilistic Approach}
\author{Barak Pinkovich, Boaz Matalon, Ehud Rivlin and Hector Rotstein
\thanks{This work was partially supported by Hyundai NGV.}
\thanks{B. Pinkovich and E. Rivlin are with the Faculty of Computer Science, Technion Israel Institute of Technology, 
        {\tt\small barakp@campus.technion.ac.il, ehudr@cs.technion.ac.il}}%
\thanks{B. Matalon is with Rafael Advanced Defense Systems Ltd.,
        {\tt\small boazm@rafael.co.il}}%
\thanks{H. Rotstein is with the Faculty of Computer Science, Technion Israel Institute of Technology and with Rafael Advanced Defense Systems Ltd.,
        {\tt\small hector@technion.ac.il}}%
        
}

\begin{document}
\maketitle
\thispagestyle{empty}
\pagestyle{empty}
\begin{abstract}

This paper considers the problem of finding a landing spot for a drone in a dense urban environment. The conflicting requirement of fast exploration and high resolution is solved using a \emph{multi-resolution} approach, by which visual information is collected by the drone at decreasing altitudes so that spatial resolution of the acquired images increases monotonically. A probability distribution is used to capture the uncertainty of the decision process for each terrain patch. The distributions are updated as information from different altitudes is collected. When the confidence level for one of the patches becomes larger than a pre-specified threshold, suitability for landing is declared.
One of the main building blocks of the approach is a semantic segmentation algorithm that attaches probabilities to each pixel of a single view. The decision algorithm combines these probabilities with \emph{a priori} data and previous measurements to obtain the best estimates.
Feasibility is illustrated by presenting a number of examples generated by a realistic closed-loop simulator.
\end{abstract}

\section{Introduction}\label{sec:1 Introduction}

\begin{figure}[hbt]
    
    \centering
    \includegraphics[width=0.9\columnwidth]{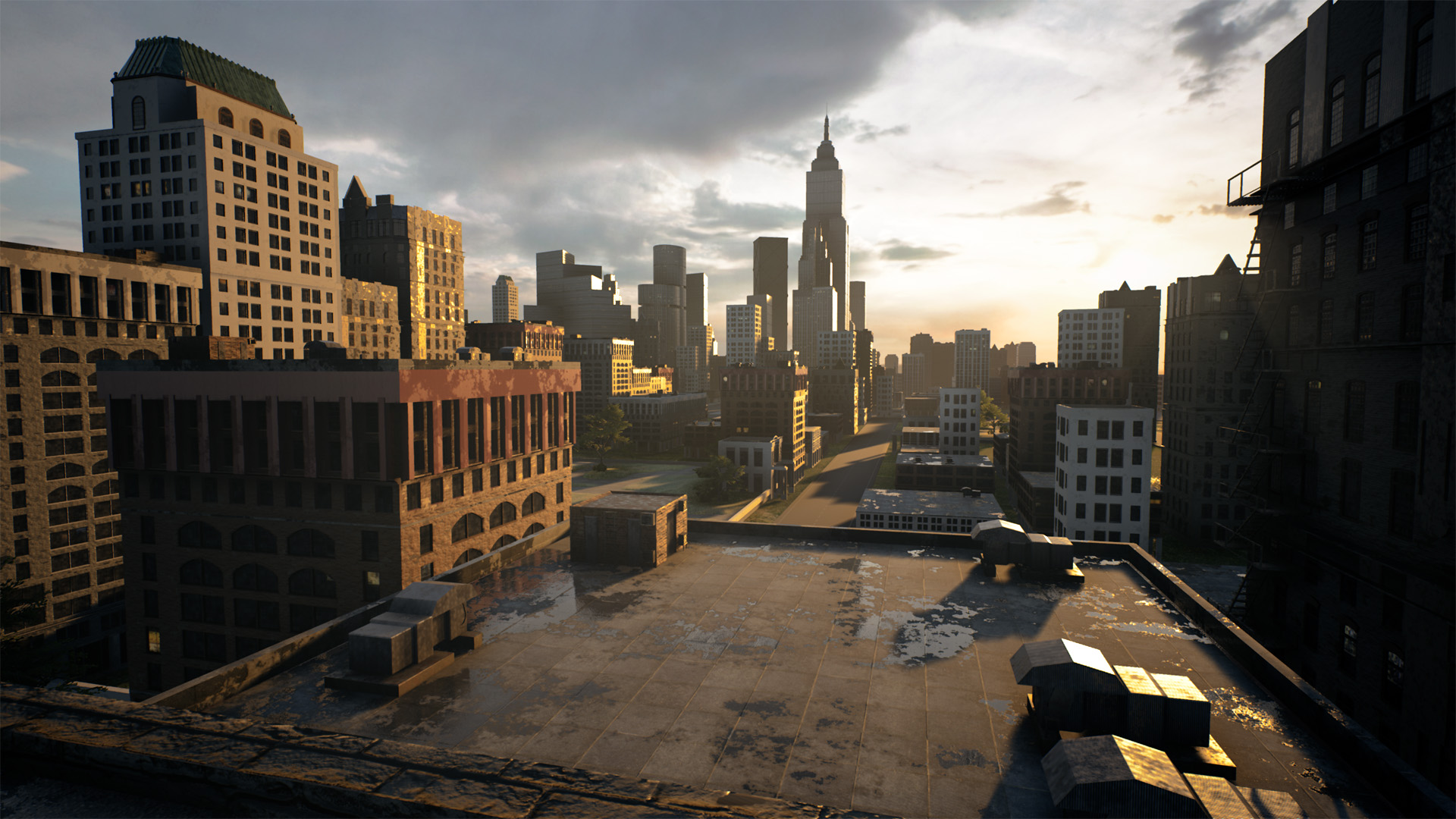}
    \caption{Simulated urban environment in which the drone attempts to land}
    \label{fig1}
\end{figure}

As opposed to conventional aircraft that take-off and land from designated and controlled areas outside city limits, future commercial drones are expected to operate smoothly in crowded urban environments. Consequently, the static and well-defined zones delimited for landing and take off will be replaced by dynamic and opportunistic areas within cities. For instance, one of the challenges of future delivery or transportation drones is to solve the "last-mile" problem, for which an autonomous drone must find a place to land having the following characteristics:
\begin{enumerate}
    \item Be relatively close to the intended destination. These places cannot be limited to pre-determined areas like heliports, sports fields, or similar. 
    \item Be appropriate for the drone size and weight.
    \item  Be appropriate for landing under harsh (or relatively harsh) flight conditions compatible with the drone's flying capabilities.
    \item Pose no safety concerns to itself, other vehicles, or living beings in the environment.
\end{enumerate}
This paper describes a multi-resolution, probabilistic approach for searching for a landing place in a dense urban environment like the one illustrated in Fig.\ref{fig1}. The approach is multi-resolution since a visual sensor from decreasing altitudes observes the urban environment, hence generating a sequence of images with monotonically increasing spatial resolution. It is also probabilistic in the sense that confidence levels of different regions of the urban region are computed based on a-priori knowledge and the result of observations.

The probabilistic viewpoint provides connections with extensive literature, including search theory and Bayes-based decision making. For example, \cite{torres2019survey} provides a survey on Bayesian networks' usage for intelligent autonomous vehicles decision-making with no focus on specific missions. Similarly, \cite{starek2016spacecraft} describes spacecraft autonomy challenges for future space missions, in which real-time autonomous decision-making and human-robotic cooperation must be considered. In a related autonomous spacecraft mission,  \cite{serrano2006bayesian} studies the selection of an appropriate landing site for an autonomous spacecraft on an exploration mission. The problem is formulated so that three main variables are defined on which to select the landing site: terrain safety, engineering factors (spacecraft's descendent trajectory, velocity, and available fuel), and the site pre-selected by using available a-priori information. The approach was tested by using a dynamics and spacecraft simulator for entry, descent, and landing.

The problem considered here is also somewhat related with \emph{forced landing}. This is because a UAV may need to decide the most suitable forced landing sites autonomously, usually from a list of known candidates  \cite{coombes2016site}. In that work, references were made to the specifications for a forced landing system laid out in a NASA technical report essentially consisting of three main criteria: risk to the civilian population, reachability, and probability of a safe landing. The emphasis is on public safety, where human life and property are more important than the UAV airframe and payload. Specifications were included in a Multi-Criteria Decision Making (MCDM) Bayesian network. See  \cite{ding2016initial} for an application of the model to a real-life example.
Initial designs for UAVs' autonomous decision systems to select emergency landing sites in a vehicle fault scenario are also considered in cite{ding2016initial}. The overall design consists of two main components: pre-planning and real-time optimization. In the pre-planning component, the system uses offline information such as geographical and population data to generate landing loss maps over the operating environment. In the real-time component, onboard sensor data is used to update a probabilistic risk assessment for potential landing areas.

Another related field of interest is that of \emph{search and rescue}, a challenging task as it usually involves a large variety of scenarios that require a high level of autonomy and versatile decision-making capabilities. A formal framework casting the search problem as a decision between hypotheses using current knowledge was introduced in \cite{chung2007decision}. The search task was defined as follows: given a detector model (i.e., detection error probabilities for false alarms and miss detections) and the prior belief that the target exists in the search region, determine the evolution of the belief that the target is present in the search region, as a function of the observations made until time $t$. The belief evolution is computed via a recursive Bayesian expression that provides a compact and efficient way to update the belief function at every time step as the searcher observes a sequence of unexplored and/or previously visited cells in the search region.
After generating a method for computing the belief evolution for a sequence of imperfect observations, the authors investigate the search control policy\textbackslash strategy.
In the context of an area search problem, \cite{bertuccelli2005robust} investigates the uncertainty associated with a typical search problem to answer a crucial question:  how many image frames would be required by a camera onboard a vehicle to classify a target as "detected" or "undetected" with uncertain prior information on the target's existence. The paper presents a formulation that incorporates uncertainty using Beta distributions to create robust search actions. As shown below,  these ideas are highly related to our approach.
\section{Problem Formulation}\label{sec:2 Problem Formulation}

Suppose a drone needs to find a landing place in an urban area $\cal{A}$. For simplicity, consider $\cal{A}$ to be planar, with an attached coordinate system $\{W\}$ such that $\cal{A}$ lies on the $x-y$ plane, measuring the altitude along the $z$-direction. Buildings and other constructions are modeled as occupied volumes over $\cal{A}$. As an example, the area considered in this paper will be 1 Km by 1 Km square. The drone can fly at different altitudes $h$ while collecting measurements using a monocular camera and a visual sensor with range capabilities. Examples of the latter are an RGB-D sensor, a Lidar, or a couple of stereo cameras. Let ${\cal{A}}_h$ be the plane parallel to $\cal{A}$ at an altitude $h$ onto which $\cal{A}$'s relevant characteristics can be mapped. For instance, a no-fly zone $\cal U$ in $\cal{A}$ (e.g., the base of a building ) will be mapped onto the corresponding subset ${\cal{U}}_h$ in ${\cal{A}}_h$ (at least if $h$ is smaller than the corresponding building's altitude). 

\begin{figure}[hbt]
    \centering
    \includegraphics[width=0.9\columnwidth]{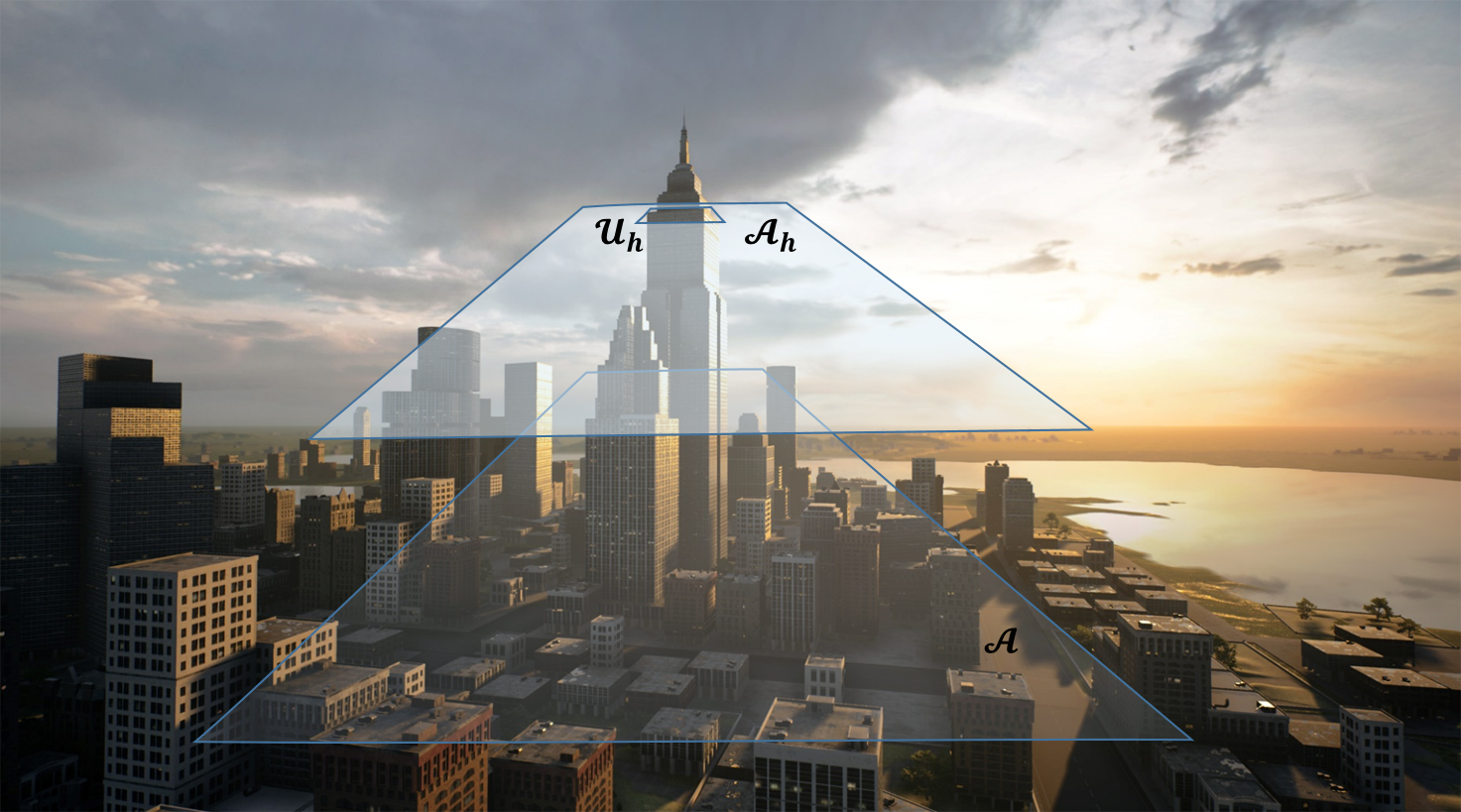}
    \caption{The area of interest $\cal{A}$, the plane ${\cal A}_h$ at an altitude $h$ and the mapping of an obstacle ${\cal U}_h$}
    \label{fig2}
\end{figure}

The mechanical structure of the drone, the size of its propellers, and the flying conditions (e.g., wind) constrain the minimum dimensions of the landing site on which the drone can safely land. Conservatively, $\cal{A}$ will be divided into a grid of identical cells $c_{ij}$, so that $\cup_{ij} c_{ij} = \cal{A}$, and each $c_{ij}$ is in principle a landing cite candidate. Note that this division is in-line with search theory (see, e.g., \cite{stone1976theory}) but stems from a different motivation: it is not a unit area being explored but the smallest area of interest. As discussed next, this will impact our developments in several ways.

The images taken by a camera on the drone at time instant $t$ will be a function of the pose $p_t$ and the camera's field-of-view. Note that the camera's orientation can differ from that of the drone by a relative rotation between the two. Assuming that the FOV is fixed and known, let $I_t(h)$ be the camera's image at time $t$ and $F_t(h)$ be the camera's corresponding footprint, namely the 3D structure mapped onto the image plane. At time $t$ the drone will have a unique pose $p_t$, but the dependence on the altitude is specifically considered in the notation; this is because the altitude scales the resolution and the footprint: for smaller $h$, one gets better resolution at the cost of a smaller footprint. The structure $F_t(h)$ is built on a collection of cells $C_t(h)=\bigcup_{\{j,k\} \in \{J_t(h),K_t(h)\}} c_{ij}\subset \cal{A}$.

In the absence of additional constraints, the drone could overfly $\cal A$ at a relatively low altitude $h_{min}$ searching for an appropriate cell $c_{ij}$ on which to land. However, at this altitude, the camera's \emph{footprint} $C_t(h_{min})$ will include a relatively small number of cells $c_{ij}$ and hence the drone would spend a potentially prohibitively amount of time/energy exploring the whole $\cal A$. On the other hand, the altitude can be selected to be the maximum allowable by regulations, say $h_{max}$, resulting in as large as possible footprints. In an extreme case, $C_t(h_{max})=\cal{A}$.  This maximizes the area subtended by a single image and minimizes the exploration time, but will give rise to an image resolution that cannot guarantee the safety of landing, e.g., will not resolve relatively small obstacles. The trade-off between altitude and resolution is solved in this work by considering a multi-resolution approach: the exploration will start at high altitude, say $h_1$, looking for the largest possible subset of $\cal{A}$ that appears to be \emph{feasible} for landing, say $L_1$. Subsequently, the drone will reduce its altitude to $h_2$ and re-explore $L_1$ with the higher resolution resulting from $h_2 < h_1$. This process results in a sequence of ${\cal A} \supset L_1 \supset \cdots \supset L_N$ that will eventually converge to a collection of one or more safe landing places.

\begin{figure}[hbt]
    \centering
    \includegraphics[width=0.9\columnwidth]{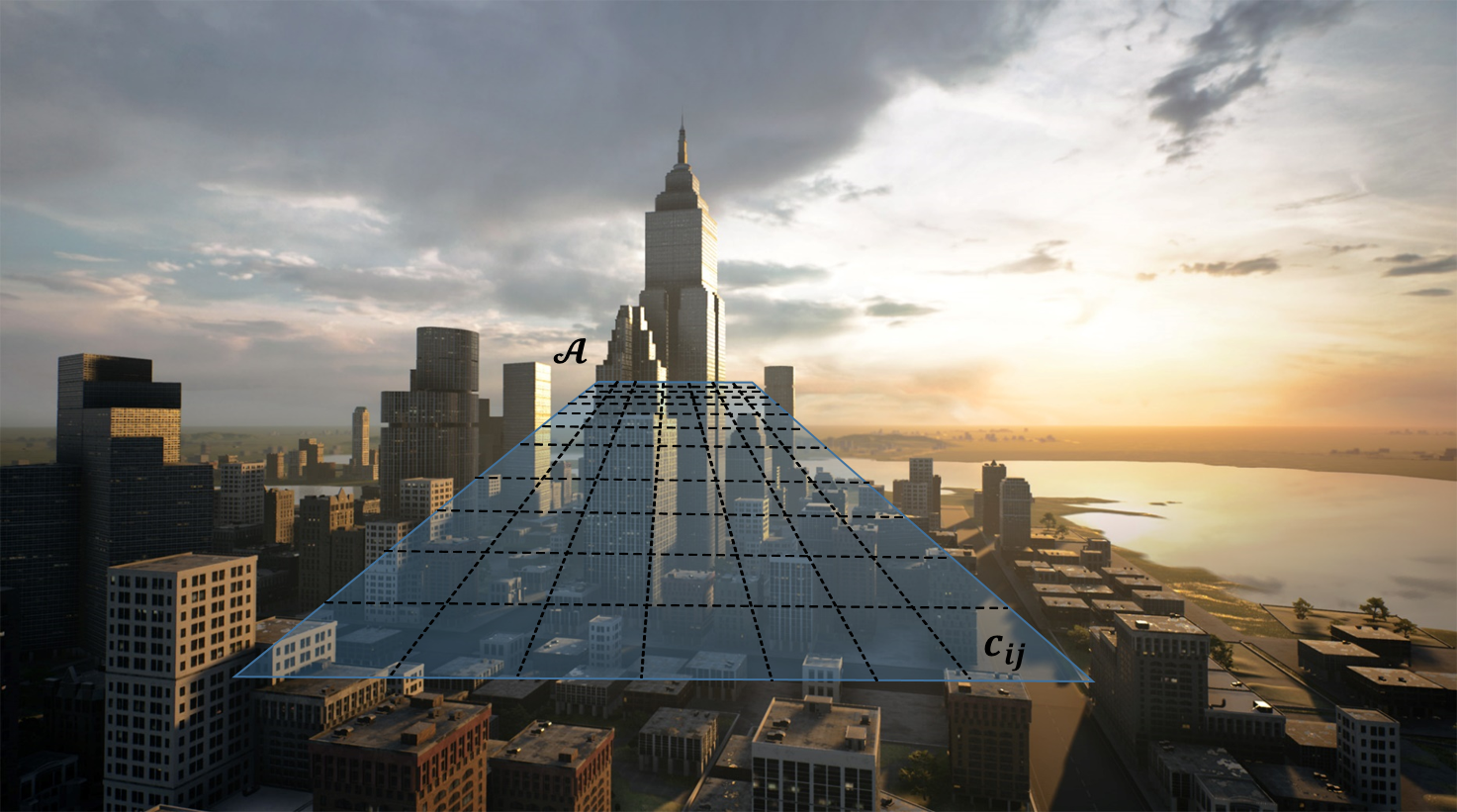}
    \caption{The plane at a given altitude is divided into small cells}
    \label{fig3}
\end{figure}
Figures \ref{fig1} to \ref{fig3} illustrate this scenario. 
\subsection{A Probabilistic Model}\label{subsec:2.2 A Probabilistic Model}

In classical search theory \cite{stone1976theory},  finding a target is often formulated as a decision problem by defining a set of binary variables:
\begin{equation} \label{eq0 binary decision porblem}
    {\cal H}_{ij}\doteq \left\{ \begin{array}{ll}
    1 & \mbox{if $ c_{ij}$ has a target} \\
    0 & \mbox{otherwise}
    \end{array}
    \right.
\end{equation}
This can be extended to the case of interest by defining:
\begin{equation} \label{eq:1 binary decision porblem}
    {\cal H}_{ij}\doteq \left\{ \begin{array}{ll}
    1 & \mbox{if $ c_{ij}$ is appropriate for landing} \\
    0 & \mbox{otherwise}
    \end{array}
    \right.
\end{equation}
In the presence of uncertainty, each cell is likely to be suitable for landing with some probability $K_{ij}$ referred to as the \emph{fitness for landing}. Clearly, if a cell is appropriate for landing, then $K_{ij}=1$, and if it is not, then $K_{ij}=0$. In real-world scenarios, prior knowledge about the fitness for landing can be based on using some kind of maps or 3D models that can be imprecise or outdated. Consequently, uncertainty in $K_{ij}$ needs to be incorporated in the model. Probably the simplest way to model the decision problem would be to introduce a Binary distribution and say that the probability of $K_{ij}=1$ is $p$ and $K_{ij}=0$ is $1-p$. However, as observed by \cite{bertuccelli2005robust}, this model fails to capture the \emph{uncertainty} of the information, and instead, it is preferable to use a Beta distribution for describing the prior knowledge together with its underlying uncertainty. Note that the Beta distribution and the Binomial and Bernoulli distributions form a \emph{conjugate pair}, so that if a Bernoulli distribution can model the sensor, then the observation of new data changes only the parameters of the prior, while the conjugacy property ensures that the posterior is in the same class (i.e., Beta). This is a critical property when propagating the belief on the fitness for landing in a Bayesian framework. Using these models allows simplifying the decision problem into a binary outcome as defined in Eq.~\ref{eq:1 binary decision porblem}. 

The Beta distribution is defined as,
\begin{equation} \label{eq:2 beta distribution}
    Pr\left(K_{ij} \left\vert \alpha,\beta \right. \right) = \dfrac{\Gamma(\alpha+\beta)}{\Gamma(\alpha)\Gamma(\beta)} K_{ij}^{\alpha-1}(1-K_{ij})^{\beta-1}
\end{equation}
Where $0 < K_{ij} < 1$ and $\Gamma(\alpha)$ is the gamma function defined as,
\begin{equation} \label{eq:3 gamma function}
    \Gamma(\alpha)= \int_{0}^{\infty} x^{\alpha-1}e^{{-x}} \,dx
\end{equation}
When $\alpha$ is an integer $\Gamma(\alpha)=(\alpha-1)!$ . The parameters $\alpha$ and $\beta$ can be considered prior "successes" and "failures". The Beta distribution is somewhat similar to the Binomial distribution. The main difference is that, whereas the random variable is ${\cal H}_{ij}$ and the parameter is $K_{ij}$ in the Binomial distribution, the random variable is $K_{ij}$ and the parameters are $\alpha$ and $\beta$ in the Beta distribution.

Bayes theorem is often used in search theory \cite{stone1976theory,chung2007decision, bertuccelli2005robust} to update the aggregated belief (e.g., posterior distribution), which is proportional to the \emph{likelihood} function times the prior distribution:
\begin{equation}\label{eq:4 Bayes posterior}
     Pr\left(K_{ij} \left\vert S_{ij}^N, \alpha, \beta \right. \right) \propto Pr\left( S_{ij}^N \left\vert , K_{ij} \right. \right) Pr\left(K_{ij} \left\vert \alpha,\beta \right. \right)
\end{equation}
Here $S_{ij}^N=\sum_{n=1}^{N}{\cal H}_{ij}^n$ is the number of successes in $N$ Bernoulli trials.  $Pr\left(K_{ij} \left\vert S_{ij}^N, \alpha, \beta \right. \right)$ is the posterior distribution for $K_{ij}$ given the number of successes. $Pr\left( S_{ij}^N \left\vert K_{ij} \right. \right)$ is the \emph{likelihood function} and $Pr\left(K_{ij} \left\vert \alpha,\beta \right. \right)$ is the prior distribution for $K_{ij}$. In \cite{chung2007decision} the likelihood distribution is given by a Binomial distribution series of $N$ observations:
\begin{equation}\label{eq:5 Binomial distribution}
    Pr\left( S_{ij}^N \left\vert K_{ij} \right. \right)=\binom{S_{ij}^N}{N}K_{ij}^{S_{ij}^N} (1-K_{ij})^{N-S_{ij}^N}
\end{equation}
Note that the underlying assumption is that the Bernoulli distribution provides an appropriate statistical model for the sensor used. This assumption is more or less natural when considering a series of noisy images. In our case, simple image processing algorithms are replaced by a more complex \emph{meta}-sensor: fitness is computed by a semantic segmentation algorithm that associates for each pixel on a given image the suitability for landing on the corresponding cell on the ground. An appropriate model for this process is discussed next.
\subsection{A Correlated Detection Model} \label{subsec:2.3 A Correlated Detection Model}

The probability framework is motivated by the limitations of the sensors used for establishing whether a given cell is appropriate for landing or not. As mentioned above, the main limitations are the camera resolution, the environmental conditions limiting visibility, and possibly scene dynamics. The probability estimated by a sensor that a given cell is appropriate for landing may vary according to altitude, with lower altitudes having higher confidence levels\footnote{Up to certain point depending on the sensor.}.

In \cite{bertuccelli2005robust} the authors assumed independent Bernoulli trials when detecting a target in recurrent visits. Independent trials are acceptable when the condition of the experiment does not change. However, in this research, our multi-resolution approach implies that when observing cell $c_{ij}$ at different altitude levels, one cannot assume uncorrelated measurements between different levels. At each level, the experiment's condition changes (e.g., different resolution), and if a landing place exists, then it is expected that the rate of success will depend on previous trials and will increase when the level of details increases while descending toward cell $c_{ij}$.

Generalizing the Binomial distribution typically involves modifying either the assumption of constant "success" probability and/or the assumption of independence between trials in the underlying Bernoulli process. The approach to generalizing the Binomial distribution in this research follows the Generalized Bernoulli Distribution (GBD) model \cite{drezner1993generalized} by relaxing the assumption of independence between trials. The GBD model was further considered in statistics literature  \cite{drezner2006limit,james2008limit,wu2012asymptotics,zhang2017limit}) with the aim to obtain its central limit theorems, including the strong law of large numbers and the law of the iterated logarithm for partial sums.

Consider a Bernoulli process $\{{\cal H}_{ij}^n,\ n\geq 1 \}$ in which the random variables ${\cal H}_{ij}^n$ are correlated so that the success probability for the trial conditional on all the previous trials depends on the total number of successes achieved to that point. More precisely, for some $0<K_{ij}<1$,
\begin{equation}
    Pr\left( {\cal H}_{ij}^{n+1} \left\vert {\cal F}_{ij}^{n} \right. \right)=(1-\theta_{ij}^{n})K_{ij} + \theta_{ij}^{n} \frac{S_{ij}^n}{n}
\end{equation}
Where $0\leq\theta_{ij}^{n}\leq1$ are dependence parameters, $S_{ij}^N=\sum_{n=1}^{N}{\cal H}_{ij}^n$ for $N\geq1$ and ${\cal F}_{ij}^{N}=\sigma({\cal H}_{ij}^{1}, \cdots,\ {\cal H}_{ij}^{N})$. If ${\cal H}_{ij}^{1}$ has a Bernoulli distribution with parameter $K_{ij}$, it follows that ${\cal H}_{ij}^{1},\ {\cal H}_{ij}^{2}, \cdots$ are identically distributed Bernoulli random variables.

By replacing the Binomial distribution in Eq.\ref{eq:4 Bayes posterior} with the GBD at each altitude $h_n$,  the aggregated belief that a cell $c_{ij}$ is suitable for landing given the number of successes will be proportional to the product of prior distribution and the altitude correlation-based distribution.
\section{Simulation}
In order to develop and test the probability multiple-resolution approach, a simulation environment was created using AirSim \cite{shah2018airsim}, a drones and cars simulator built on the Unreal Engine \cite{UnrealEngine} AirSim is an open-source, cross-platform simulator for physically and visually realistic simulations. It is developed as an Unreal plug-in that can be integrated into any Unreal environment.
Within AirSim, a drone can be controlled using a Python/C++ API; for our project's requirements, the drone can be configured similarly to a real drone in terms of dynamics, sensor data, and computer interface. The drone can be flown in the simulation environment from one way-point to another while acquiring data from the sensors defined in the platform. For the current configuration, the simulator computes images taken by a downwards-looking camera and a GPS/inertial navigation system.

\begin{figure}[hbt]
\begin{tabular}{cc}
    \includegraphics[width=0.21\textwidth]{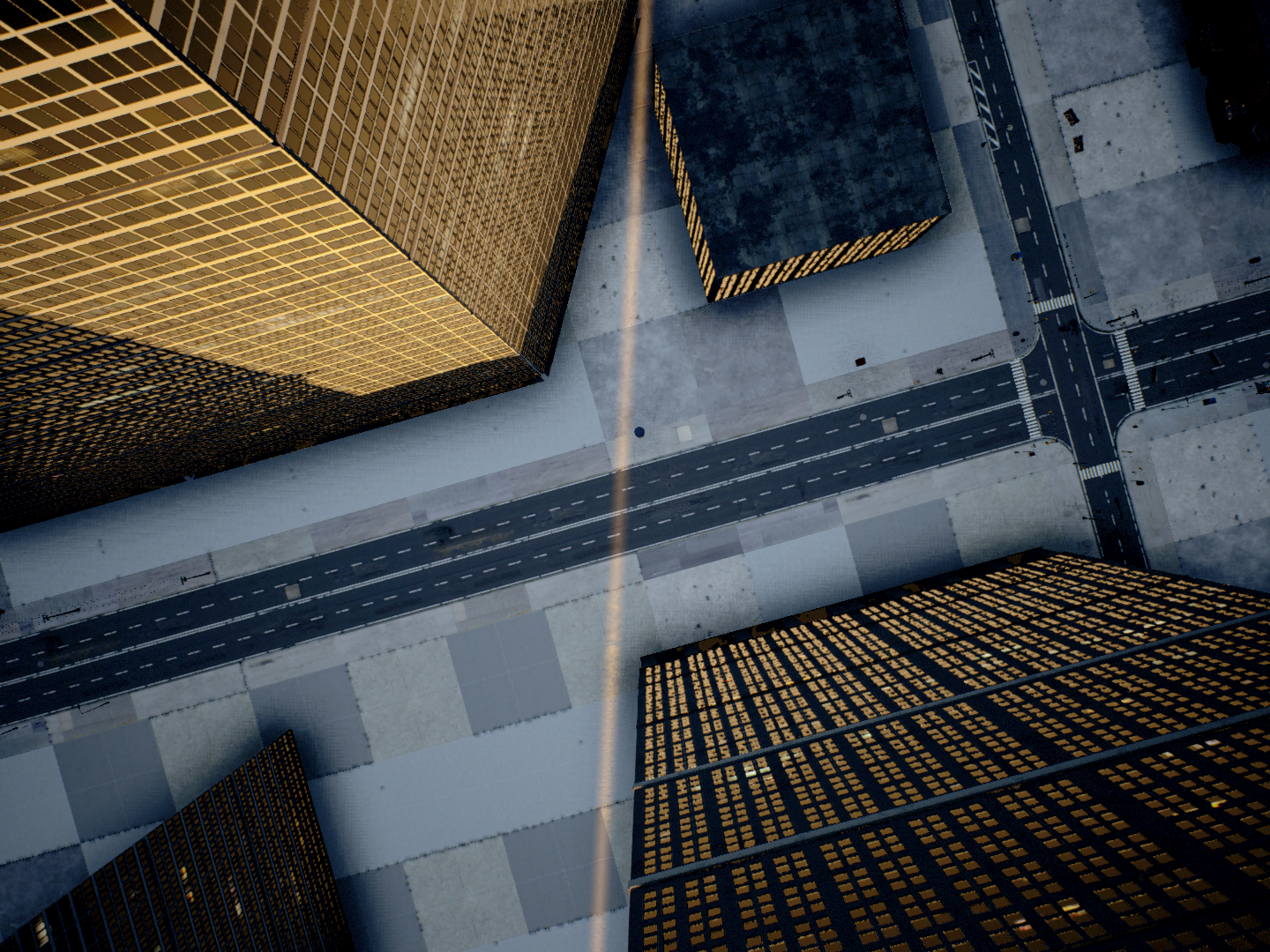} &
    \includegraphics[width=0.21\textwidth]{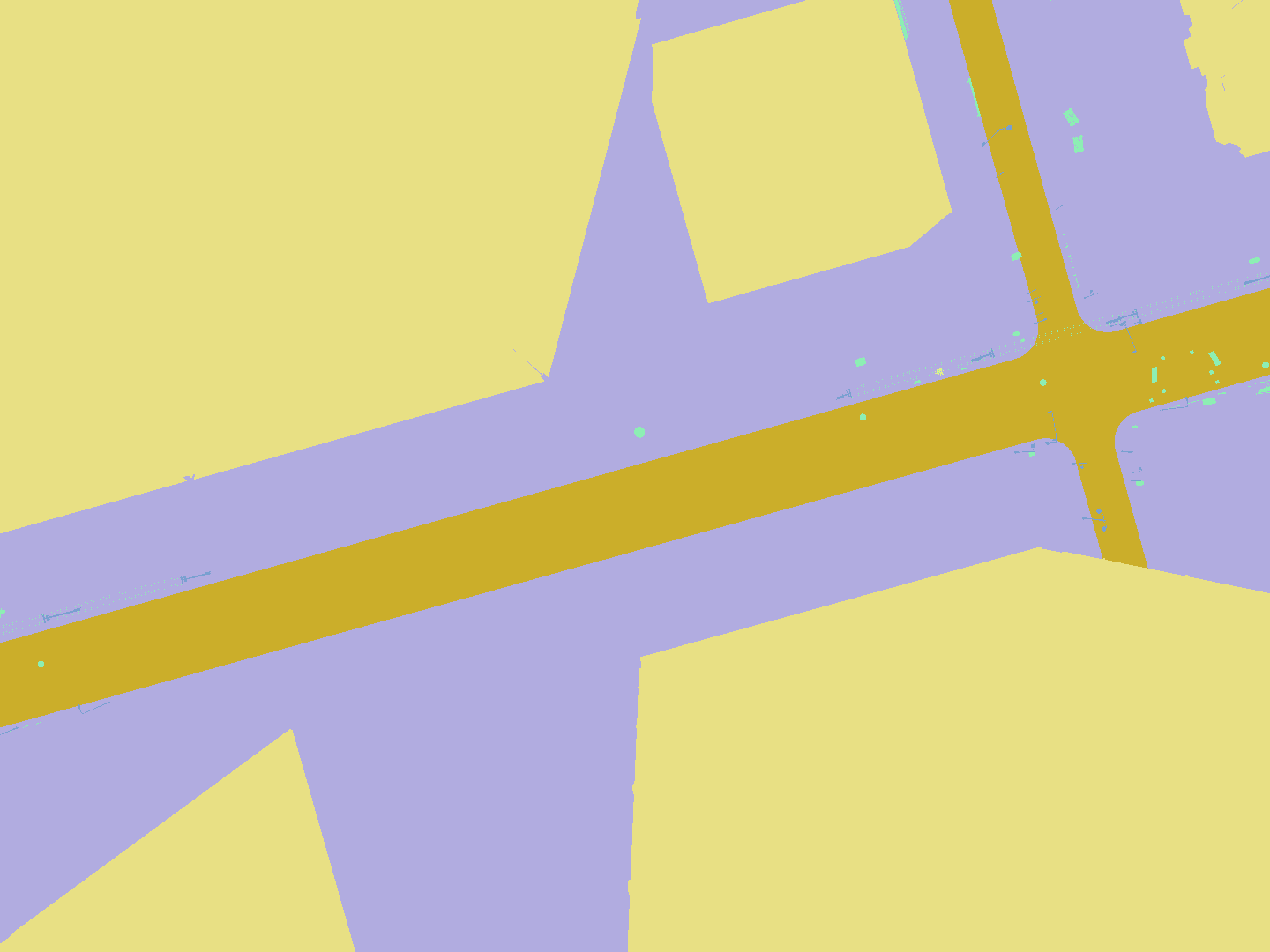}
\end{tabular}
    \caption{High altitude urban scene captured with a downward looking camera. On the left, a simulated image. On the right, the corresponding segmented scene.}
    \label{fig4}
\end{figure}

\begin{figure}[hbt]
\begin{tabular}{lr}
    \includegraphics[width=0.21\textwidth]{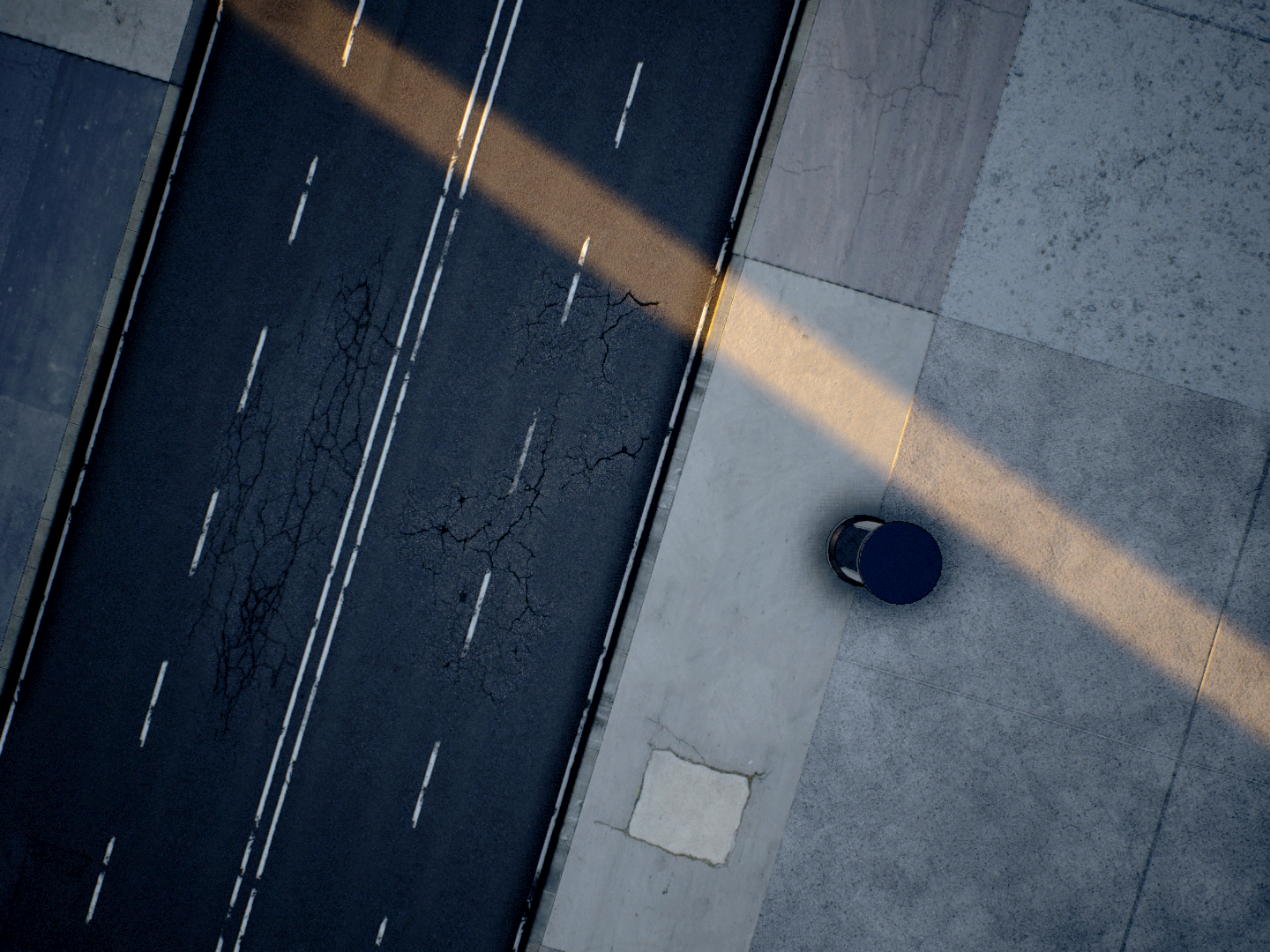} &
    \includegraphics[width=0.21\textwidth]{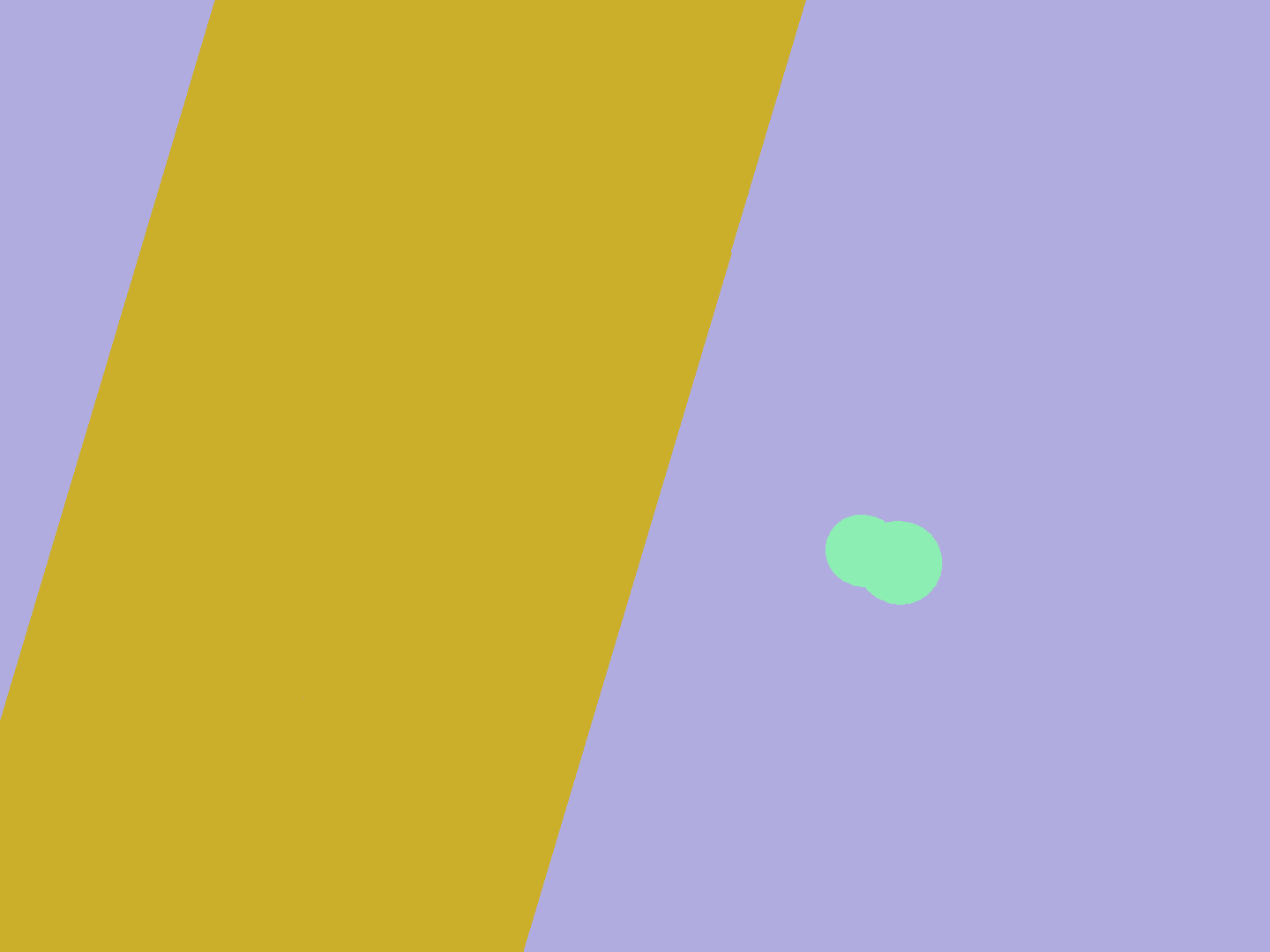}
\end{tabular}
    \caption{Low altitude urban scene captured with a downward looking camera. On the left, a simulated image. On the right, the corresponding segmented scene.}
    \label{fig5}
\end{figure}

The simulator has two main purposes:
\begin{enumerate}
    \item Test the overall multi-resolution approach as the Unit Under Test (UUT). In this mode, the simulator functions as the Hardware In The Loop (HIL) tester's data generator. The data generated by the simulator is streamed into the drone's mission computer; the system processes the data and computes the next coordinate to which the drone flies. Note that in this case, the simulator drives the real-time functioning of the closed-loop system.
    \item Generating off-line data for the search algorithm. As mentioned above, the drone can be flown using the API around the map at different scenarios and heights while generating data at pre-determined rates. Typical data consists of RGB images, segmented images, and inertial navigation data, forming a probabilistic model analysis and validation data set. 
\end{enumerate} 
The 3-D model used in the simulator was the Brushify - Urban Buildings Pack \cite{Brushify}, purchased from the Unreal marketplace.
Figures~\ref{fig4} and \ref{fig5} show a simple example of RGB and corresponding segmentation images for the cameras simulated on the drone. The images highlight the observation that objects that occupy a cell are hardly detectable from a high altitude (e.g., phone booth), while when descending, the gathered information allows the algorithm to detect these objects and decide that the drone cannot land in these specific cells. 
\section{Analysis and Preliminary Results}

Obtaining some prior knowledge about the urban scene is necessary for having a probability model. For such purpose, a labeled 3-D Digital Surface Model (DSM) was generated using the simulation by using a $2x2[m]$ cell resolution.
\begin{figure}[hbt]
    \centering
    \includegraphics[width=0.99\columnwidth]{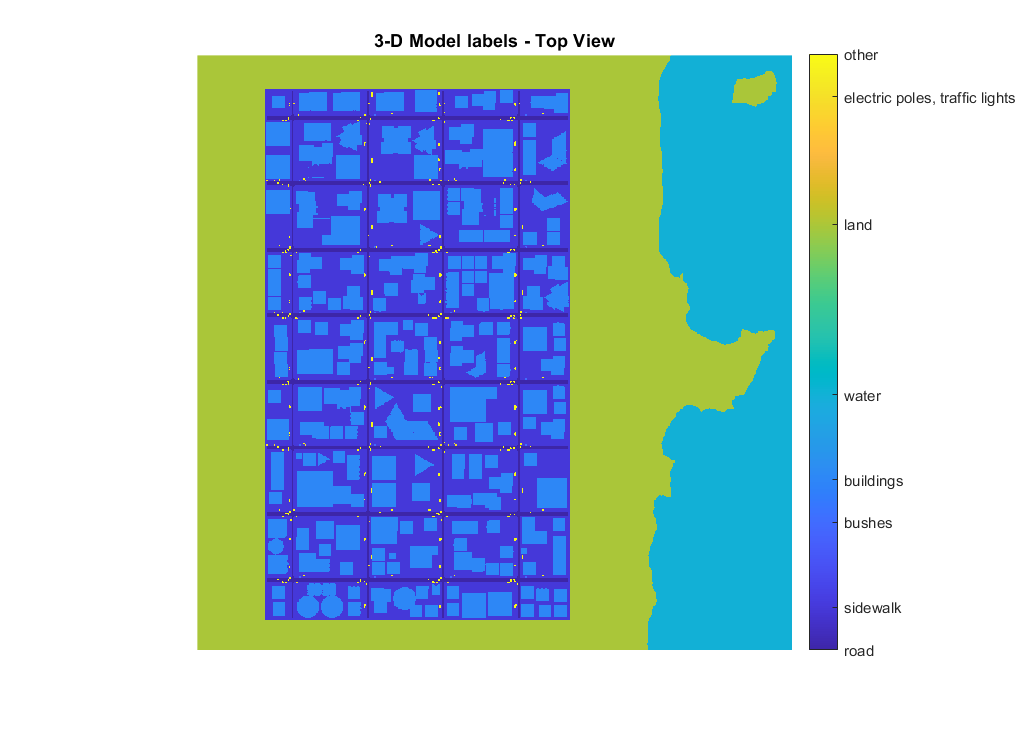}
    \caption{Top View of the Digital Surface Model with chosen labels}
    \label{fig6}
\end{figure}
The DSM, shown in Fig.\ref{fig6}, allows choosing the parameters of the prior distribution for each cell given the label of that cell. The labels that were chosen to be represented with initial probabilities were such that they were visible from a high altitude and may be considered appropriate for landing or not when descending. Fig.\ref{fig7} shows the Beta distribution's parameters for each label. These parameters were chosen to give some knowledge on an appropriate (or not) place to land. Still, there is sufficient uncertainty in the prior's belief to allow some degree of freedom to change the values and the new belief with new observations.
\begin{figure}[hbt]
    \centering
    \includegraphics[width=0.8\columnwidth]{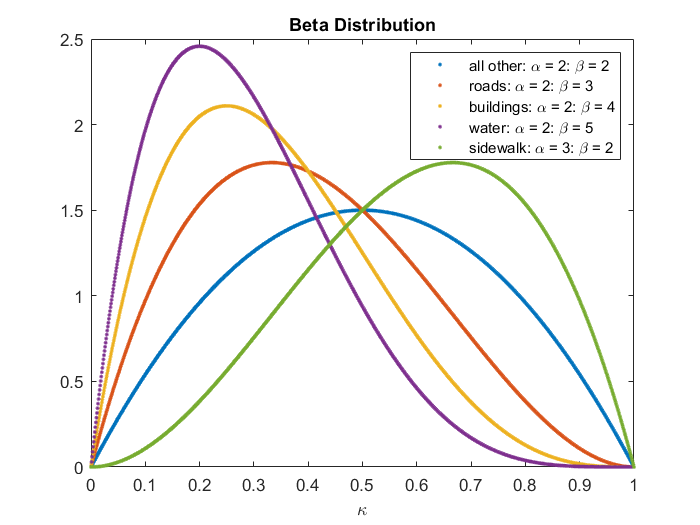}
    \caption{Prior Beta Distribution For Chosen Labels}
    \label{fig7}
\end{figure}

\begin{figure}[hbt]
\begin{tabular}{cc}
    \includegraphics[width=0.21\textwidth]{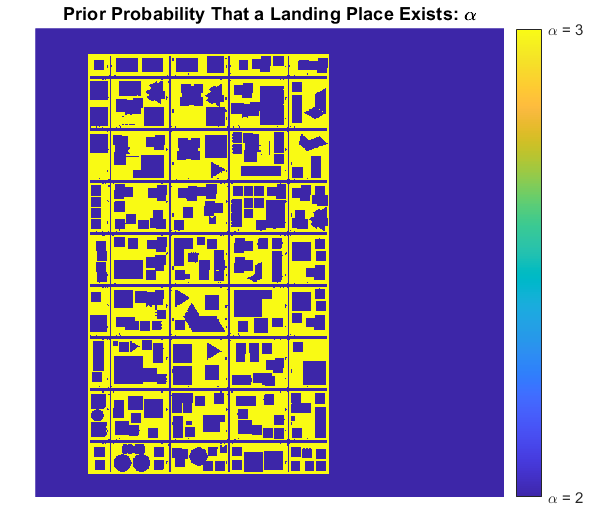} &
    \includegraphics[width=0.21\textwidth]{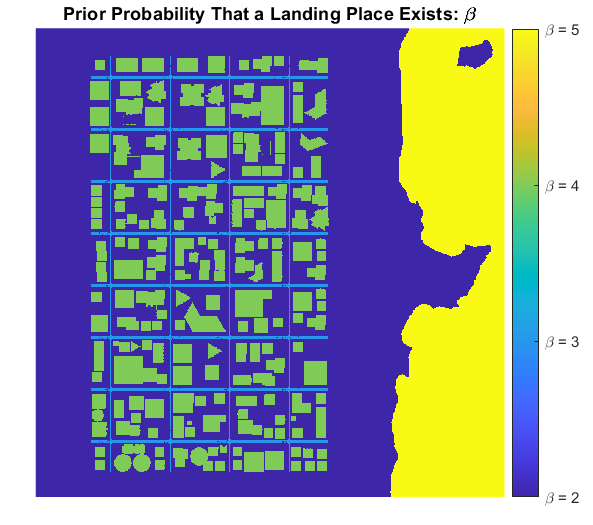}
\end{tabular}
    \caption{Prior $\alpha$ and $\beta$ parameters for each cell in the urban scene}
    \label{fig8}
\end{figure}

\begin{figure}[hbt]
    \centering
    \includegraphics[width=0.9\columnwidth]{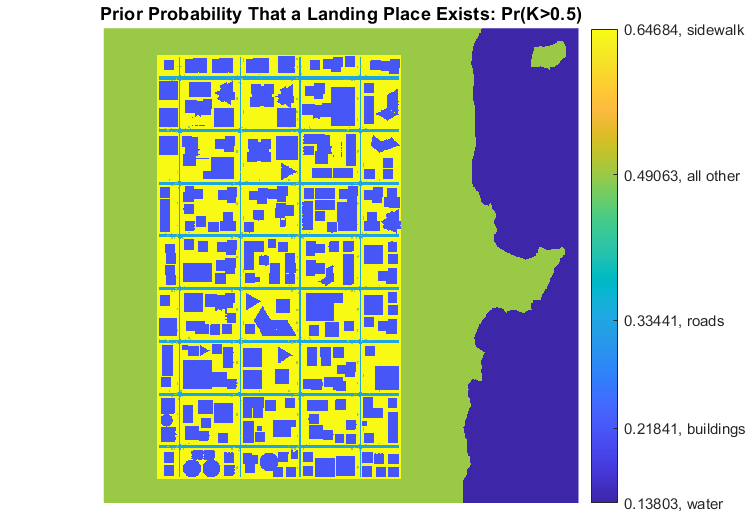}
    \caption{Prior Landing Probability For $\kappa = 0.5$}
    \label{fig9}
\end{figure}

The $\alpha$ and $\beta$ parameters in Eq.\ref{eq:2 beta distribution} are stored for each cell $c_{ij}$ in a 2-D model of the urban scene as illustrated in Fig.\ref{fig8}. To determine if cell $c_{ij}$ contains a place to land, the probability $Pr(K_{ij}>\kappa)$ is calculated: 
\begin{equation} \label{eq:6 integrating belief}
    Pr(K_{ij}>\kappa)= \int_{\kappa}^{1} Pr(K_{ij}) \,dK
\end{equation}
For instance, Fig.\ref{fig9} shows the prior belief that a landing place exists for $Pr(K_{ij}>0.5)$. Only sidewalks are somewhat appropriate for landing even for $\kappa=0.5$, and the belief can easily be changed when new observations are obtained.

The downwards-looking camera mounted on the drone provides color images that need to be converted into information on how much each cell is suitable for landing. Clearly, this relation may be highly complex. In recent years, deep neural networks have shown to be successful for various Computer-Vision applications, including the kind of Semantic Segmentation problems relevant to our purpose. Consequently, we chose to employ the semantic segmentation network  BiSeNet \cite{yu2018bisenet}, which fuses two information paths - context path and spatial path. The context allows information from distant pixels to affect a pixel's classification at the cost of reduced spatial resolution.
In contrast, the spatial path maintains fine details by limiting the number of down-sampling operations. This net also provides a reasonable compromise between segmentation accuracy and compute requirements. Other network architectures, potentially more complex and accurate, will be explored in future work.

The network was trained and validated on images taken by the camera while flying in the urban environment at different altitudes. The model uses the labels in Fig.\ref{fig6} for training, and during inference, the model predicts probability scores (summing to 1) for the different categories. Each category is also assigned with an a-priori weight, representing how much this category is suitable for landing (e.g. weight[sidewalk] = 0.8 and weight[building] = 0). To obtain a final score $p_{mn}$ for each image pixel, we take a weighted average of the categories` probabilities using the pre-defined weights. $p_{mn}$, which can vary between $0$ to $1$, describes the probability that a landing site exists based on the observed data. Using the 6-DOF of the drone, the image footprint, i.e., pixels coordinates projected on the ground, were transformed to a world coordinate system to be associated with each cell $c_{ij}$ in the grid. The outcome of a Bernoulli trial for success or failure is given by counting $N_{p}$, the number of pixels associated to cell $c_{ij}$ that pass $p_{mn}>0.5$ and are relative to $N_{pc}$, the total number of pixels associated with $c_{ij}$. If the relative amount is greater than $0.99$, then $c_{ij}$ holds a successful trial.
\begin{equation} \label{eq:8 relative number of pixels greater than 0.5}
    N_{p} = \sum_{m,n \in i,j}^{N_{pc}}\{p_{mn}>0.5\}
\end{equation}
\begin{equation} \label{eq:9 Bernoulli pixel trial}
    {\cal H}_{ij}\doteq \left\{ \begin{array}{ll}
    1, & \mbox{ $\dfrac{N_{p}}{N_{pc}}>0.99$} \\
    0, & \mbox{otherwise}
    \end{array}
    \right.
\end{equation}
\subsection{A Single Altitude Bayesian Update}\label{subsec:4.1 Single Altitude Bayesian Update}

\begin{figure}[hbt]
\begin{tabular}{cc}
    \includegraphics[width=0.21\textwidth]{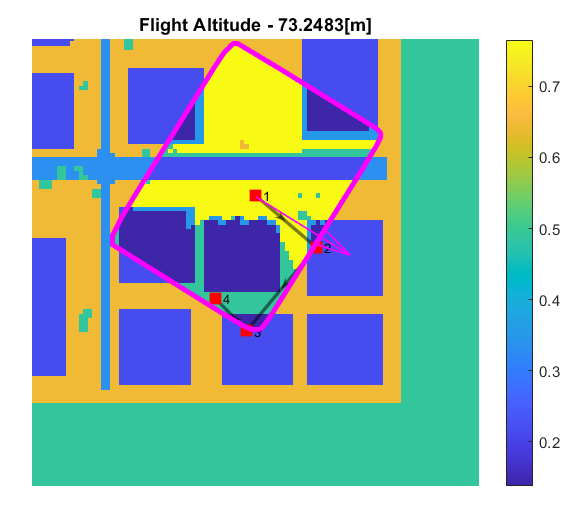} &
    \includegraphics[width=0.21\textwidth]{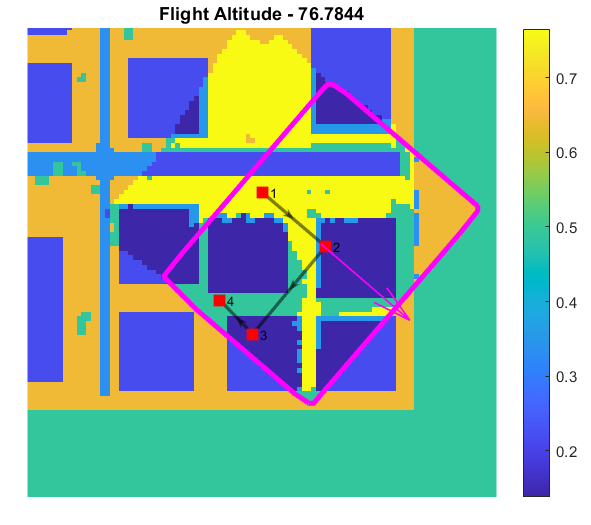}
\end{tabular}
    \caption{Bernoulli trial update - each cell is updated only once}
    \label{fig10}
\end{figure}

\begin{figure}[hbt]
\begin{tabular}{lr}
    \includegraphics[width=0.21\textwidth]{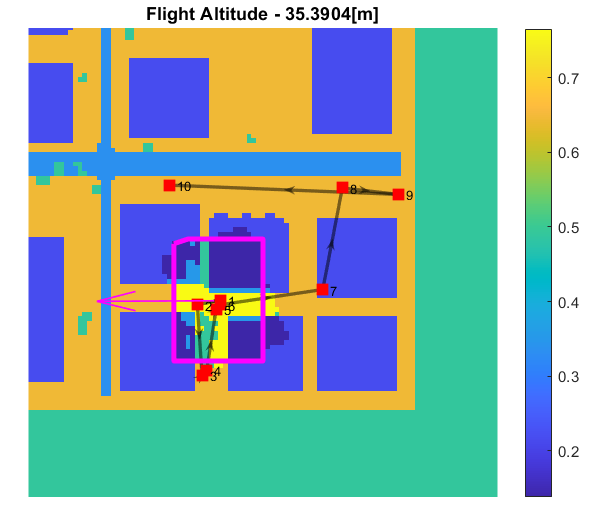} &
    \includegraphics[width=0.21\textwidth]{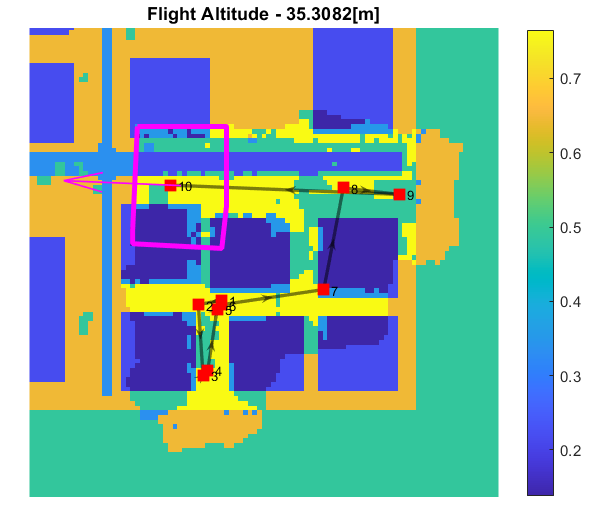}
\end{tabular}
    \caption{Bernoulli trial update - each cell is updated only once}
    \label{fig11}
\end{figure}

The Bayesian update was tested for several flight scenarios.Suppose now that the drone flies and takes images at a constant altitude and that the semantic segmentation network analyzes images. Each cell $c_{ij}$ belonging to an image footprint is associated with the corresponding projected pixels, and a Bernoulli trial is performed according to Eq.\ref{eq:9 Bernoulli pixel trial}. The trial is performed on each cell only once to prevent added correlation effects at that altitude. The outcome of the single-trial is added to the $\alpha$, and $\beta$ values previously selected as the prior (shown in Fig.\ref{fig7}, \ref{fig8}) and then integrated numerically according to Eq.\ref{eq:6 integrating belief}. Figures~\ref{fig10} and \ref{fig11} show the updating stage at different time instances and altitudes.

\subsection{ An Altitude-Based Bernoulli Trials Distribution}\label{subsec:4.1 Altitude-Based Bernoulli Trials}

\begin{figure}[hbt]
\begin{tabular}{cc}
    \includegraphics[width=0.21\textwidth]{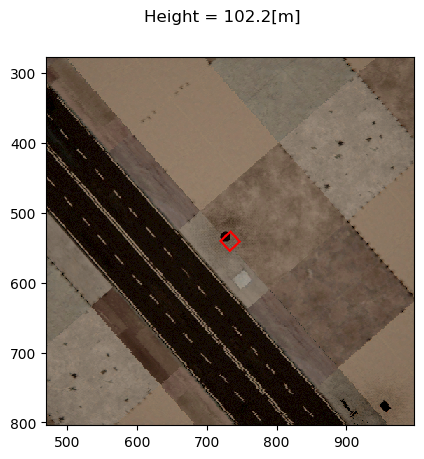} &
    \includegraphics[width=0.21\textwidth]{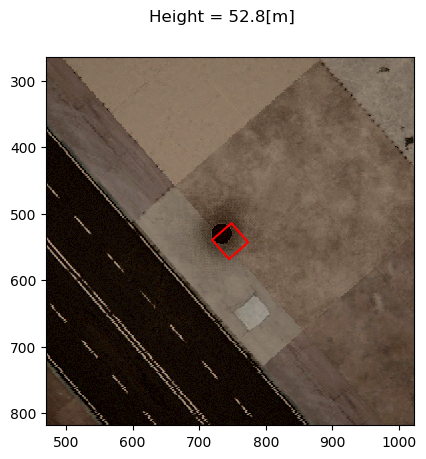}
\end{tabular}
    \caption{Semantic segmentation input - telephone booth partially occupies the cell}
    \label{fig12}
\end{figure}

\begin{figure}[hbt]
\begin{tabular}{lr}
    \includegraphics[width=0.21\textwidth]{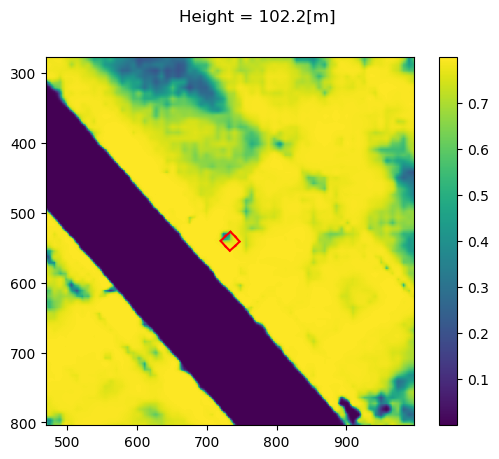} &
    \includegraphics[width=0.21\textwidth]{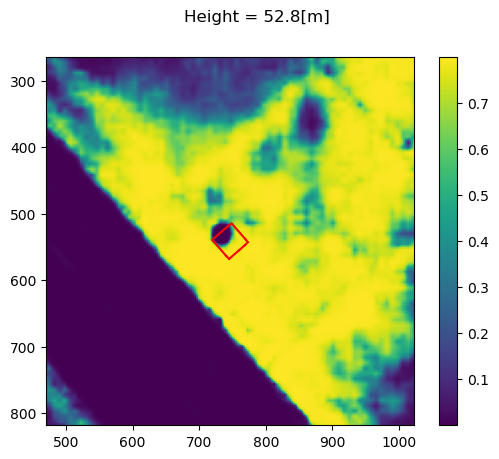}
\end{tabular}
    \caption{Semantic segmentation output - telephone booth partially occupies the cell }
    \label{fig13}
\end{figure}

In order to study the GBD model, an experiment with Bernoulli trials at different altitudes was performed. There are several objects placed on the ground at different locations around the urban scene. The experiment was planned so that a single object is selected for the drone to descend upon at different locations. At each location, images are taken as input for the semantic segmentation network. On each output of the network, a Bernoulli trial is performed on a single cell according to Eq.\ref{eq:9 Bernoulli pixel trial}, so that at each altitude that an image is taken, there is a single success or failure output on a given cell $c_{ij}$. Fig.\ref{fig12}, \ref{fig13} show the input and output at selected altitudes in a single location. A phone booth was selected for the drone to descend upon. There are $31$ locations around the urban scene with the phone booth placed on the sidewalk. A cell in the world coordinate system was selected at each location so that the phone booth occupies partially or the entire cell. The expected outcome would be that the phone booth will be partially detected at high altitudes, and the selected cell would be detected as fit for landing. In contrast, more details will be detected when descending, and the cell will be detected as unfit for landing.

\begin{figure}[hbt]
    \centering
    \includegraphics[width=0.8\columnwidth]{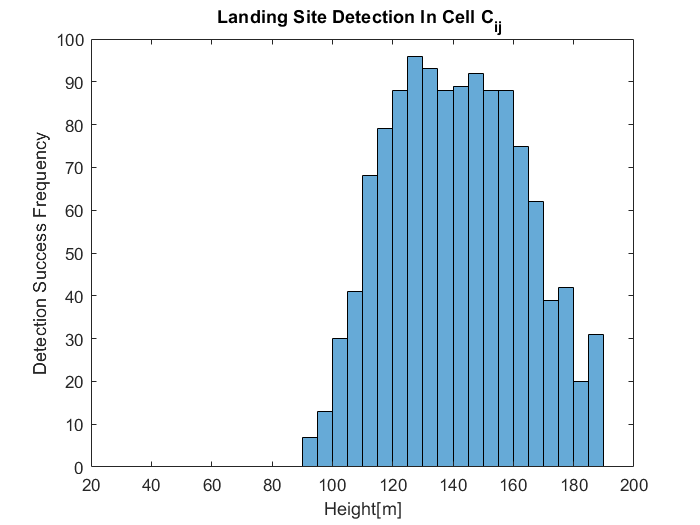}
    \caption{an Altitude-Based Bernoulli Trials Histogram}
    \label{fig14}
\end{figure}

Fig.\ref{fig14} shows the histogram for the $31$ locations with altitude-based trials. We can see that the aforementioned expected behavior is indeed observed under 160 meters approximately. Peculiarly, above this altitude, the success frequency diminishes markedly. This may be explained by the fact that there were no images from these altitudes in the training set, making the net's prediction unreliable. It should be noted that even in lower altitudes, its prediction can be noisy due to various reasons, particularly the training set being not diverse enough. We expect that more diversity in the training set will yield greater reliability of the net, which in turn will result in a better fit for the model.


\section{Conclusions and Further Work}

This paper has presented a multi-resolution probabilistic approach for finding an appropriate landing site for a drone in a dense urban environment. The approach uses a-priori data (e.g., a map or a DSM) to estimate the fitness for landing probability distribution for each cell on which the environment is divided. Distribution and not probabilities are used in an attempt to model the uncertainty of the data. Subsequently, the data collected by a visual sensor and processed by a semantic segmentation neural net is used to update the distribution using Bayesian networks. In order to do that, the probability of success is factored into the results obtained by the net. Images are captured at different altitudes in an attempt to solve the trade-off between image quality, including spatial resolution, context, and others. After presenting theoretical aspects, the simulation environment in which the approach was developed and tested is detailed, and the experiments conducted for validation are described. The overall approach is shown to produce the desired results, at least for the simulation environment in which it was tested.

Further work is currently underway in three main directions. Firstly, we would like to establish some success criteria for the procedure. For instance, we would like to develop bounds to enable more accurate ways of analyzing our results . Secondly, we would like to generate more realistic images on which the semantic segmentation can be trained and tested. Lastly, we would like to test the approach on actual data and conduct a flight-test to achieve real-life validation.
 \section*{Acknowledgement}
 We would like to thank Distinguished Professor Daniel Wehis, head of the Technion Autonomous System Program (TASP) for his continuous support. 
\bibliographystyle{ieeetr}
\bibliography{IEEEabrv,Refs.bib}{}

\end{document}